\documentclass[runningheads]{llncs}
\usepackage{graphicx}
\usepackage{times}
\usepackage{caption}
\usepackage{xurl}
\usepackage{subcaption}
\usepackage{amsmath}
\usepackage{amsfonts,amssymb}
\usepackage[ruled, linesnumbered]{algorithm2e}

\begin{document}
\title{CAGE: Causality-Aware Shapley Value \\ for Global Explanations}
\author{Nils Ole Breuer\inst{1}\and
Andreas Sauter\inst{2}\and
Majid Mohammadi\inst{2} \and
Erman Acar\inst{3}}
\authorrunning{Breuer et al.}
% First names are abbreviated in the running head.
% If there are more than two authors, 'et al.' is used.
%
\institute{GT-ARC gGmbH, Germany\\ \email{nils.breuer@gt-arc.com}\\ \and
Vrije Universiteit Amsterdam, Amsterdam, The Netherlands\\ \email{\{a.sauter,m.mohammadi\}@vu.nl}\\ \and
ILLC and IvI, University of Amsterdam, Amsterdam, The Netherlands\\\email{erman.acar@uva.nl}}

\maketitle

\begin{abstract}%DUMMY  ABSTRACT, WE NEED a PROPER one
  %This paper proposes a causally-aware global explanation framework for predictive models based on Shapley values. We introduce a novel sampling procedure for out-of-coalition features that respects the causal relations of input features. We derive a method that incorporates causal knowledge into global explanation and offers the possibility to interpret the predictive feature importance considering their causal relation. We also evaluate our method on synthetic datasets and on a real-world application. Results of synthetic experiments show that the causal explanation method is closer to the real contribution of the features to the target.%
As Artificial Intelligence (AI) is having more influence on our everyday lives, it becomes important that AI-based decisions are transparent and explainable. As a consequence, the field of eXplainable AI (or XAI) has become popular in recent years. One way to explain AI models is to elucidate the predictive importance of the input features for the AI model in general, also referred to as global explanations. Inspired by cooperative game theory, Shapley values offer a convenient way for quantifying the feature importance as explanations. However many methods based on Shapley values are built on the assumption of feature independence and often overlook causal relations of the features which could impact their importance for the ML model. Inspired by studies of explanations at the local level, we propose CAGE (\textbf{C}ausally-\textbf{A}ware Shapley Values for \textbf{G}lobal \textbf{E}xplanations). 
%on both local and global level. Recently, there have been extensions towards regarding causal relations between features on the local level.  However, at a global explanation level such considerations are still missing. This leads to problematic assumptions for the sampling procedure which lies at the heart of the Shapley value-based methods. This paper aims to fill this gap, by proposing a causally-aware global explanation framework for predictive models based on Shapley values. 
In particular, we introduce a novel sampling procedure for out-coalition features that respects the causal relations of the input features. We derive a practical approach that incorporates causal knowledge into global explanation and offers the possibility to interpret the predictive feature importance considering their causal relation. We evaluate our method on synthetic data and real-world data. The explanations from our approach suggest that they are not only more intuitive but also more faithful compared to previous global explanation methods.
\end{abstract}


\section{Introduction}
Explainable artificial intelligence (XAI) is a field of study in artificial intelligence (AI) research that complements complex machine learning (ML) models with comprehensible insights, facilitating human understanding and trust \cite{langer2021we}. It endeavors to unravel the intricate decision-making processes of AI systems, providing clear, interpretable, and accessible explanations that favorably align with human cognition and reasoning. As AI technologies burgeon and permeate various sectors —from healthcare \cite{paul2021artificial} and finance \cite{wang2021survey} to criminal justice \cite{zavrvsnik2021algorithmic}—  the imperative for XAI is magnified, demanding that the opaque ``black-box" models are more transparent and the algorithmic decisions thereof are justified.

In this ever-evolving landscape, various research lines and methodologies have been proposed, each aspiring to shed light on different aspects of the workings of AI models \cite{saeed2023explainable}. Among these, Shapley value-based methods such as SHAP \cite{lundberg2017unified} and SAGE \cite{covert2020feature} have gained substantial traction, employing concepts from cooperative game theory to attribute locally (instance-based) and globally (model-based) the contribution of each feature to a model’s prediction, respectively. Despite their popularity (in particular the former), these methods, built on assumptions of feature independence, often overlook the nuanced causal relationships and interactions amongst features, potentially leading to oversimplified or misleading explanations.

Pivotal work \cite{miller2019explanation} underscores that genuine explanations are intrinsically tied to causality, reflecting a philosophical viewpoint where explanations are crafted through counterfactual reasoning — envisaging alternative scenarios and assessing their impact on outcomes. Thus, causality emerges as a fundamental pillar in crafting meaningful and intuitive explanations \cite{miller2019explanation,mittelstadt2019explaining}. In such a pursuit, recent explorations such as  Causal SHAP \cite{heskes2020causal} and Asymmetric SHAP \cite{frye2020asymmetric}  have sought to infuse causality into local explanation frameworks, as they emerge as promising frontiers, endeavoring to intertwine causal reasoning with \textit{local} explanation techniques. These methods are argued to reflect the human cognitive processes of causal inference, striving for explanations that resonate with innate human understanding and intuition while being grounded in strong mathematical foundations \cite{miller2019explanation}.

%In our study, we explore the complex relationship between causality and \textit{global} explanations. We investigate the potential of global explanation methods that incorporate causal information and examine how they can enhance the trustworthiness of insights obtained from an ML model. We show that by incorporating concepts of causality we can improve the deficiencies mentioned above and build more faithful global explanations. 

In this article, we introduce a method that incorporates a causal lens into Shapley-value based \textit{global} explanations (i.e., SAGE), abbreviated by the acronym CAGE. Empowered by its capability to express complex causal relations between features, we show both theoretically and empirically that CAGE can alleviate the aforementioned deficiencies, and result in more faithful global explanations. More specifically, the main contributions of this article are as follows. 
\begin{enumerate}
    \item We introduce a model-agnostic causality-aware conceptual framework based on Shapley values for the global explanations i.e., CAGE. In particular, we establish a novel sampling procedure that respects the causal relations of input features;

    \item We theoretically show that CAGE satisfies desirable causal properties; an indication that it is designed from first principles. 
    
    \item We carry out an empirical analysis with both synthetic and real-world data, concluding that explanations resulting from CAGE are more faithful compared to their causally agnostic counterparts.
\end{enumerate}

In the remainder of this paper, we start by introducing core concepts and our notation. We then present CAGE in detail, show that it derives causally sound explanations, and present an algorithm to estimate its values. Furthermore, we apply CAGE to synthetic and real-world data to substantiate our claims. Finally, we discuss the most related works and provide an in-depth discussion about our results before we conclude. The code for our framework and experiments is available at \url{https://anonymous.4open.science/r/CausalGlobalExplanation-2DB1}.

%%%%%%%%%%%%%%%%%%%%%%%%%%%%%%%%%%
%%% Preliminaries and Notation
%%%%%%%%%%%%%%%%%%%%%%%%%%%%%%%%%%
\section{Preliminaries and Notation}
In this section, we introduce the key concepts and notations used throughout this paper. These are additive importance measures, Shapley values, and fundamental notions of causality.
%In the previous Section, we superficially introduced some of the methods and concepts of the field of explainable artificial intelligence research. In this Section, we will elaborate and define some of these concepts that are important to the goal of this study in more detail.

\subsection{Causal Models and Interventions} \label{causality}
\subsubsection{Structural Causal Models (SCM)} We resort to \emph{structural causal models (SCM)} to express causal relations formally. An SCM is a tuple $M = (\mathbf{X}, \mathbf{F}, \mathbf{U}, \mathbf{P})$ of observed variables $\mathbf{X} = \{X_1, \ldots, X_n\}$, unobserved exogenous variables $\mathbf{U} = \{U_1,\ldots , U_n\}$, functional relations $\mathbf{F}$ that define direct causal effects, and $\mathbf{P}$ a set of pairwise independent distributions of exogenous variables. Each SCM induces a directed acyclic graph (DAG), where the direct causes of an endogenous variable are incoming edges (parents). Formally, each variable $X_i$ is determined by  $f_i \in \mathbf{F}$ s.t. $X_i \leftarrow f_i(\mathbf{Pa_i}, U_i)$, where $\mathbf{Pa_i} \subseteq \mathbf{X}\setminus \{X_i\}$ (parents) are the direct causes of $X_i$. %Furthermore, each $U_i$ is distributed according to $P_i \in \mathcal{P}$.
%To insert the causal perspective in global explainable artificial intelligence we resort to two concepts of causality. The first one is that of structural causal models (SCM) with which dependencies between variables can be specified. In an SCM each variable is generated, in this work, by a deterministic function of other variables in the SCM. An SCM entails a directed acyclic graph (DAG) and 
From the conditional independence assumption a joint probability distribution $P(X_1, ..., X_n) = \prod_{i=1}^n P(X_i | \mathbf{Pa_i})$ of the SCM can be inferred \cite{pearl1995causal,scholkopf2021toward,scholkopf2022causality} which we will refer to as \emph{observational distribution}. %Formally, each variable $X_i$ is determined by a function $f$ such that:

%\begin{equation}
%    X_i := f(\mathbf{Pa_i}, U_i),
%\end{equation}

%where $\mathbf{Pa_i}$ refers to the parents of $X_i$ in a graphical sense, or the cause from a purely causal perspective, and $U_i$ are mutually independent noise terms. 

\subsubsection{Causal Chain Graphs} 
\begin{figure}
    \centering
    \includegraphics[width=0.6\textwidth]{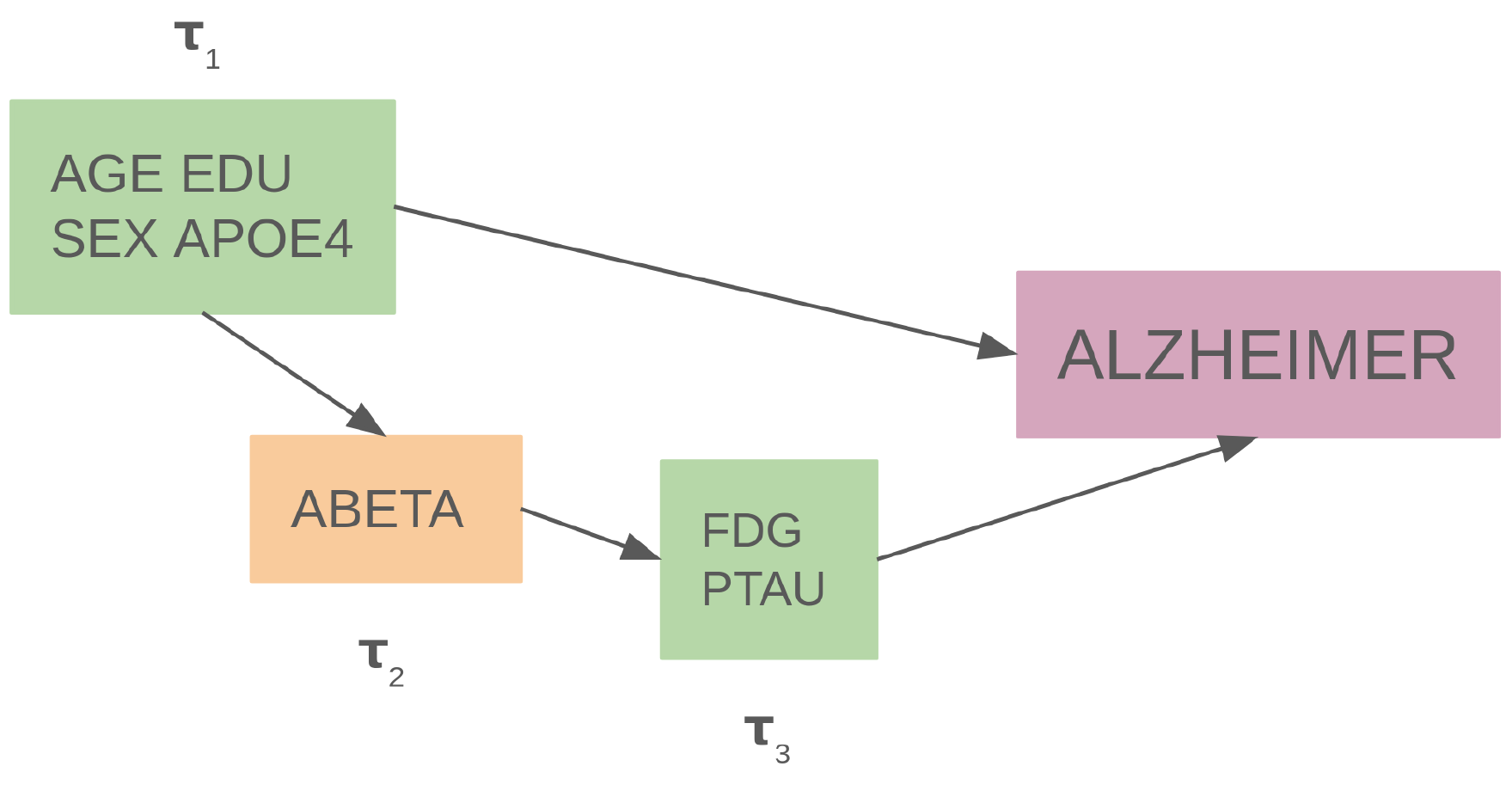}
    \caption{Causal chain graph that shows the partial causal ordering of the Alzheimer dataset \textit{ADNI} \cite{jack2008alzheimer} used later in the experiments. The chain graph consists of three components $\tau_1$, $\tau_2$, and $\tau_3$ that have a causal ordering. In the components, the complete causal relationships of the variables are not known. Variables in $\tau_1$ and $\tau_3$ are assumed to have common confounders (green) and $\tau_2$ is assumed to have causal interaction (yellow). The target variable is marked in red.}
    \label{fig:causalchaingraph}
\end{figure}

We use the notion of causal chain graphs \cite{lauritzen1989graphical} as a relaxation on the information encoded in causal graphs. Causal chain graphs represent a partial causal ordering of sets of variables (chain components) among which the causal relationships are not fully known. This means that variables are in the same component whenever they have a common confounder or mutual interaction between them (see  Fig. \ref{fig:causalchaingraph} for an example).

\subsubsection{Interventions} Interventions are purposefully modifying the values of variables to discern cause-and-effect relationships. Therefore, they are of vital importance for causal reasoning. An intervention is formally expressed by the so-called \emph{do-operator}, denoted for $\mathbf{Y} \subseteq \mathbf{X}$ as $do(\mathbf{Y} = \mathbf{c})$ or $do(\mathbf{Y})$ when the value is clear from the context. This forces one or more variables in an SCM to a particular value, effectively replacing all corresponding functions with this value. Graphically, this corresponds to pruning all incoming edges to that variable. Interventions result in a new joint distribution $P(\mathbf{X} \setminus \mathbf{Y}\mid do(\mathbf{Y}= \mathbf{c}))$ 
\cite{pearl2012calculus} which we will refer to as the \emph{post-interventional distribution}.
%\begin{equation}
%    P_M(Y | do(X_i := c)) = P_{M_c}(Y)
%\end{equation}
If there are changes between the observational and post-interventional distributions, conclusions can be drawn about the causal influence of the intervention variable on the other variables \cite{pearl1995causal,peters2017elements}. 
%We shall use interventions to change the conditional (or marginal) distribution that is inherent in Equation (\ref{sage_value}) to an interventional distribution. 
%Thus, consider causal relations and effects in between features so that the explanations and assigned feature importances respect these relations. 

\subsection{Shapley Additive Global Importance}\label{sec:SAGE}
%Additive importance measures are a crucial concept in the domain of model agnostic explainability, enabling us to dissect the contributions of individual features in AI models. The feature importance determined by these methods can be interpreted as how much 
In the domain of XAI, \emph{additive importance measures} are of common practice since they provide model-agnostic explainability by dissecting contributions of individual features in the target AI models.  Intuitively, feature importance can be interpreted as the amount of 
%Covert et al. \cite{covert2020understanding} introduced the class of additive importance measures which combines methods that are built to compute feature importance. The authors interpret feature importance as how much 
predictive power a feature provides for an AI model at hand.
%Formally, additive importance measures are methods that assign scores $\phi$ to features $i \in N$ so that the additive function $u(S)$ represents the predictive power of a feature subset $S$ as the difference to a constant $\phi_0$ as shown in Equation \ref{aim} \cite{covert2020understanding}. %, i.e. when a feature is important then adding that feature will give a better prediction. Formally \cite{covert2020understanding} define the class of additive importance measures as:
%\begin{definition}[Additive importance measures \cite{covert2020understanding}] Additive importance measures are methods that assign scores $\phi$ to features $i \in N$ so that the additive function $u(S)$ represents the predictive power of a feature subset $S$ as the difference to a constant $\phi_0$.
%\begin{equation} \label{aim}
%    u(S) = \phi_0 + \sum_{i \in S} \phi_i
%\end{equation}
%\end{definition}
%Methods in this class sum up the values $\phi_i$ for each feature.
The main characteristic of these measures is that the individual feature importances $\phi_i(v)$ sum up to the models' overall prediction power w.r.t. a value function $v$ \cite{covert2020understanding}. Whenever $v$ is clear from the context, we will only write $\phi_i$. %$\phi_i$ can be interpreted as the increase in performance of the model prediction, if feature $X_i \in D \setminus S$ is added to $S$ 

%The additive importance measures class is strongly related to the class of additive feature attribution methods that were introduced in the paper of Lundberg and Lee \cite{lundberg2017unified}. The difference between the two classes is how the importance of a feature is defined. In the paper of Covert et al. \cite{covert2020understanding} the importance is equal to the increase in model performance whereas  the importance of a feature in \cite{lundberg2017unified} is defined by how much it influences the prediction, i.e. how much does the output of the model differ from the average output. Lundberg and Lee \cite{lundberg2017unified} have shown that the concept of \textit{Shapley values} satisfies desirable properties associated with additive methods. This is based on the way \textit{Shapley values} are built. 
A broadly used instance of additive measures are the so-called \emph{Shapley values}. In the context of XAI, the Shapley value of a feature $i$ measures the marginal increase of the value function $v$ of a model if that feature is considered in the prediction. The Shapley value of each feature $i$, $\phi_i$ is then the weighted sum of that increase over all possible permutations of all features $D$ of a specific subset of features $S$. This is inspired by cooperative game theory and defined by \cite{shapley1953value} as follows:

\begin{equation} \label{shapley}
    \phi_i(v) = \frac{1}{|D|} \sum_{S  \subseteq D \setminus \{i\}} \binom{ |D|-1}{|S|}^{-1} \left( v(S \cup {i}) - v(S) \right),
\end{equation}
%In Section \ref{related_work}, it was already mentioned that \textit{Shapley values} originated from the field of game theory where the function $v$ is defined as a cooperative game. Translated to XAI the value function $v$ is associated with the output of a model. Equation (\ref{shapley}) measures the increase of the value function $v$ if a feature $i$ is added to a subset $S$. $\phi_i$ is then the weighted sum of that difference over all possible permutations of all features $D$ of the feature subset $S$. This means that the value function $v$ only gets a subset of features but because we want to analyze the model that was trained on all features $D$, a way to sample the missing features $\bar{S}$ needs to be devised.
where $D$ is the set of all feature indices, $S$ is a subset of $D$ also called the ``in-coalition" features, and $i$ is the specific feature indices which is added to $S$ and for which the importance is computed. The value function $v$ only gets a subset of feature indices, but because we want to analyze the model that takes all features $D$ as input, a way to sample the values of the missing feature indices $\bar{S} = D \setminus S$ needs to be devised.

To derive \emph{Shapley additive global importance (SAGE)} values for a model, Covert et al.  \cite{covert2020understanding}  defines a value function that measures the change in the loss $\mathcal{L}$ of a model, shown in Equation \ref{sage_value}:
\begin{equation} \label{sage_value}
    v_f(S) = -\mathbb{E}[\mathcal{L}(\mathbb{E}[f(\mathbf{X}) | \mathbf{X}_S], Y)],
\end{equation}
%Equation \ref{sage_value} shows how the interpretation of feature importance as predictive power gets measured.

\noindent where $\mathbf{X}_S$ are the variables for the ``in-coalition" features $S$, $\mathbf{X} = \mathbf{X}_S \cup \mathbf{X}_{\bar{S}}$ are all feature variables, and $Y$ is the prediction target. To compute the prediction $f(\mathbf{X})$ one must first sample $\mathbf{X}_{\bar{S}}$ of the ``out-coalition" features $\bar{S}$. The standard way \cite{lundberg2017unified} to do this is to sample $\mathbf{X}_{\bar{S}}$ from a conditional distribution $P(\mathbf{X}_{\bar{S}} | \mathbf{X}_S = \mathbf{x}_S)$, where $\mathbf{x}_S$ are the realized values of $\mathbf{X}_S$. This is done in the inner expected value where we marginalize the out-coalition features $\bar{S}$.  This sampling procedure assumes feature independence which can lead to to spurious explanations and misrepresenting feature dependencies \cite{kumar2020problems}. $v_f(S)$, therefore calculates the prediction quality if only the values of the features $S$ are known by averaging over the features $\Bar{S}$ and this as an average over an entire data set. For the average over the entire data set the outer expected value is used. In Section \ref{sec:methodology} we describe how we overcome these shortcomings, i.e., the independence assumption, by considering causal models of the data.

%where $v_f$ computes the expected loss with respect to the real target values $Y$ of the prediction $f(X)$ given $X_S$, i.e., the values of the features in $S$ are known and the values of the missing features $X_{\bar{S}}$ in $\bar{S}$ must be sampled. The standard way to sample the missing features The inner expected value marginalizes out the missing features $\bar{S}$ conditioned on the known features in $S$ from the distribution $P(X_{\bar{S}} | X_S = x_S)$. This conditional sampling method for the missing features $\bar{S}$ works under the assumption of feature independence which can lead %Several papers have pointed this out, see Section \ref{related_work}, to lead 
%to spurious explanations and misrepresenting feature dependencies \cite{kumar2020problems}. The outer expected value estimates $v_f$ for the whole dataset X as we are calculating global importance values.  In Section \ref{sec:methodology} we describe how we overcome these shortcomings by considering causal models of the data.

\section{Causality-Aware Global Explanations}\label{sec:methodology}
In this section, we introduce our causality-aware global importance measure.
%Moreover, we explain what our method measures and how this is a new interpretation. 
In addition, we show that this measure has some desirable (causal) properties (cf. Theorem~1). And last, we provide an approximation algorithm for computing it. 

%and the practical implementation of our method. 

% In this Section, we explain the details of how we compute a causality-aware global importance measure and sketch the practical implementation of our method.

%As has been stated above the goal of this paper is to build a framework to incorporate causal knowledge into global explanations. In the previous Section, we defined and elaborated important concepts for our causal explanation framework. Now we will introduce our framework and give a practical implementation for it.

\subsection{Global Causal Shapley Values} \label{sec:globalcausalshap}
%Global causal Shapley values \eqref{sage_value}, as explained above, rely on the conditional distribution $P(X_{\bar{S}} | X_S = x_S)$ to handle missing features.
%If we revisit the value function (\ref{sage_value}) of the global explanation method \textit{SAGE} \cite{covert2020understanding} we see that it takes the expected value of a conditional distribution $P(X_{\bar{S}} | X_S = x_S)$ to handle the missing features. 
%This means that missing features %(out-of-coalition features) 
%are sampled by a conditional distribution given the known features %(in-coalition features)
%. In practice, assuming the independence of the features, this distribution is exchanged with a marginal distribution $P(X_{\bar{S}})$. Considering the complexity of many real-world systems, it is unlikely that the independence assumption holds in general, hence it can lead to spurious explanations \cite{kumar2020problems}. % We will incorporate the concepts of causality that have been introduced in Section \ref{causality} and will define a causal value function for the class of additive importance measures methods. 

Considering the complexity of many real-world systems, it is unlikely that the independence assumption of \cite{covert2020feature} in the global explanation methods holds in general, hence it can lead to spurious explanations \cite{kumar2020problems}. 
For that reason, we propose a global explanation method that considers causal dependencies when sampling out-coalition features, following the recent works on computing causality-inspired feature importance for local explanations \cite{heskes2020causal,janzing2020feature,jung2022measuring}. For our sampling procedure, we assume a causal graph to be given. More specifically, we sample the out-coalition features $\mathbf{X}_{\bar{S}}$  from a post-interventional distribution (after intervening on the known features of interest $\mathbf{X}_S$) instead of a conditional distribution i.e.,  $P(\mathbf{X}_{\bar{S}} | do(\mathbf{X}_S = \mathbf{X}_S))$ instead of  $P(\mathbf{X}_{\bar{S}} | \mathbf{X}_S= \mathbf{X}_S)$.  This leads us to a sampling procedure from the post-interventional distribution, resulting in the following causal value function:

%To this end we will use interventions \cite{pearl1995causal, pearl2012calculus} to change the conditional distribution to an interventional distribution \cite{heskes2020causal, jung2022measuring, janzing2020feature}. We define the value function as follows:

\begin{equation} \label{causal_v}
    v_f(S) = -\mathbb{E}_{\mathbf{X}Y}[\mathcal{L}(\mathbb{E}_{\mathbf{X}_{\bar{S}}}[f(\mathbf{X}) | do(\mathbf{X}_S = \mathbf{x}_S)], Y)],
\end{equation}
which is determined by marginalizing the out-coalition features $\bar{S}$ from the post-interventional distribution \cite{heskes2020causal}:
\begin{equation} \label{marginalization}
   \mathbb{E}[f(\mathbf{X}) | do(\mathbf{X}_S = \mathbf{X}_S)] = \int f(\mathbf{X}_{\bar{S}}, \mathbf{X}_S) P(\mathbf{X}_{\bar{S}} | do(\mathbf{X}_S = \mathbf{x}_S)) d\mathbf{X}_{\bar{S}}.
\end{equation}
%In other words, we will change the conditioning by observation in Equation (\ref{sage_value}) to conditioning by intervention. We intervene on the in-coalition features $S$ and sample the out-of-coalition features $\bar{S}$ from a causal interventional distribution with the do-operator.
Through this intervention and by marginalizing the out-coalition features, we ensure the independence of the in-coalition features. % we take the causal relations between the features into consideration. Therefore, overcoming the independence assumption. 
%In detail the sampling distribution changes from $P(X_{\bar{S}} | X_S)$ to $P(X_{\bar{S}} | do(X_S = x_S))$. 

\subsection{Properties of Global Causal Feature Importance}\label{sec:properties}

 Our causal feature importance measure comes with a set of  theoretical guarantees, that have been introduced in  Jung et al. \cite{jung2022measuring}. Intuitively, these are the desirable properties which ensure \emph{causal soundness} of such measure. We present them below first, and then show that they are satisfied also in our global method CAGE. 

\begin{itemize}
    \item[P1]\textbf{Perfect assignment:} The global causal contributions are perfectly assigned if  ${\mathbb{E}}[\mathbb{I}(1)] - \mathbb{E}[\mathbb{I}(0)] = \sum_{i \in D} \phi_i$ where $\mathbb{I}(1)$ and $\mathbb{I}(0)$ correspond to the loss in Equation \eqref{causal_v} with intervention on all features and no intervention, respectively. 

    \item[P2]\textbf{Causal irrelevance:} If $X_i$ is causally irrelevant to $Y$ for all $\mathbf{W} \subseteq \mathbf{X} \setminus \{X_i\}$ s.t. $\forall y,  P(y \mid  do(X_i, \mathbf{W})) = P(y \mid do(\mathbf{W})) $, then $\phi_i = 0$, i.e., if a feature does not have any causal predictive power then CAGE value is 0. 

    \item[P3] \textbf{Causal symmetry:} If $X_i, X_j \in \textbf{X}$ have the same causal contribution to the predictive power of $Y$ for all $\mathbf{W} \subseteq \mathbf{X} \setminus \{X_i, X_j\}$  s.t. $\forall y,  P(y \mid  do(X_i, \mathbf{W})) = P(y \mid do(X_j, \mathbf{W})) $, then $\phi_i = \phi_j$, i.e., if two features have the same causal predictive power then the features have the same CAGE value.

     \item[P4] \textbf{Causal approximation:} For any $S \subseteq D$: $\forall i\in S$, $\phi_i$ well approximates $\mathbb{E}[Y \mid \text{do}(\mathbf{X}_S)]$ i.e.,
     $\{\phi_i\}_{i=1}^{n} = \arg \min_{\{\phi'_i\}_{i=1}^{n}} \sum_{S \subseteq D} (\mathbb{E}[Y \mid \text{do}(\mathbf{X}_S)] - \sum_{i\in S} \phi'_i)^2 \omega(S)$ for some positive and bounded function $\omega(S)$.
\end{itemize}

Intuitively, P1  means that the sum of all causal feature contributions corresponds to the average treatment effect if we intervene on all features compared to no intervention. In particular,  it captures how each feature contributes to the predictive power. P2 ensures that if a feature does not have any causal contribution to the predictive power then it has an importance value of zero. P3  means two features with the same causal predictive power have the same importance values.  Posing the importance values as the solution to a weighted least square problem in P4 ensures that we can consider them as approximations of the causal effect.

\begin{theorem} CAGE is causally sound i.e., the derived values have properties P1 to P4.\end{theorem}

\begin{proof}
Let the value function $v_f$ be defined as in Equation \eqref{causal_v}. 
%\begin{equation}
%    v_f(S) = {\mathbb{E}}[- l({\mathbb{E}}[f(X) | do(X_S = x_s)], Y)]
%\end{equation}

To show perfect assignment (P1) i.e., $\sum_{i \in D} \phi_i = \mathbb{E}[\mathbb{I}(1)] - \mathbb{E}[\mathbb{I}(0)]$, following \cite{vstrumbelj2014explaining} we can write $$\phi_i(v_f) = \frac{1}{\mid D\mid !} \sum_{\pi \in \Pi(D)} \{v_f(\{i\} \cup Pre_\pi(i)) - v_f(Pre_\pi(i))\}$$ where $\pi$ is a permutation from the set of all possible permutations of feature indices $D$ and $pre_\pi(i)$ is the predecessor of $i$ in the permutation $\pi$. By summing all feature contributions we get
    \begin{align*}
        %\phi_i(v_f) &= \frac{1}{n!} \sum_{\pi \in \Pi([n])} \{v_f(i \cup Pre_\pi(i)) - v_f(Pre_\pi(i))\}\\
        \sum_{i=1}^{\mid D \mid}\phi_i(v_f) &= \frac{1}{\mid D \mid!} \sum_{\pi \in \Pi(D)} \sum_{i=1}^{\mid D \mid} \{v_f(\{i\} \cup Pre_\pi(i)) - v_f(Pre_\pi(i))\}\\
        &= \frac{1}{\mid D \mid!} \sum_{\pi \in \Pi(D)} \{v_f(D) - v_f(\emptyset)\}\\
        &= v_f(D) - v_f(\emptyset)\\
        &= \mathbb{E}[- \mathcal{L}(\mathbb{E}[f(\mathbf{X}) | do(\mathbf{X})], Y)] - \mathbb{E}[- \mathcal{L}(\mathbb{E}[f(\mathbf{X}) | do(\emptyset)], Y)]\\
        &= \mathbb{E}[\mathbb{I}(1)] - \mathbb{E}[\mathbb{I}(0)]
    \end{align*}
    \noindent which shows the equality in P1.
    %\noindent where we define $\mathbb{I}(1)$ as the loss between the post-interventional prediction of the model and the true target $\mathbf{Y}$ if an intervention is done on all features \textbf{X}. $\mathbb{I}(0)$ is defined as the loss with no intervention on the features, respectively.

   To show causal irrelevance (P2), we assume $X_i$ to have no causal contribution to the prediction of $Y$ for all $S \subseteq D \setminus \{i\}$. Then, according to Equation \eqref{causal_v}:
   \begin{align*}
       v_f(S \cup \{i\}) &= -\mathbb{E}[\mathcal{L}(\mathbb{E}[f(\mathbf{X}) | do(\mathbf{X}_S = \mathbf{x}_S, X_i = x_i)], Y)]\\
       &= -\mathbb{E}[\mathcal{L}(\mathbb{E}[f(\mathbf{X}) | do(\mathbf{X}_S = \mathbf{x}_S)], Y)]\\
       &=  v_f(S)
   \end{align*}
   Hence, according to the definition of Shapley values \eqref{shapley} $\phi_i = 0$.
   %$v_f(S \cup \{X_i\}) - v_f(S) = 0$ holds for $\forall S \subseteq D \setminus i$. Resulting in desired property $v_f(S \cup \{X_i\}) = v_f(S)$.
   
    Although the proofs for P3 and P4 correspond to the ones in \cite{jung2022measuring}, we include them here for completeness and to provide an easy map to our notation. To show causal symmetry (P3) we assume the features $X_i$ and $X_j$ have the same causal predictive power. % with regard to all $\mathbf{W} \subseteq \mathbf{X} \setminus \{X_i, X_j\}$. This results in $v_f(\{X_i\} \cup S) = v_f(\{X_j\} \cup S)$ $\forall S \subseteq D \setminus \{i, j\}$. 
    Therefore,
    \begin{align*}
        \phi_i(v_f) &= \frac{1}{\mid D \mid} \sum_{S \subseteq D \setminus \{i\}} {\binom{\mid D \mid -1}{|S|}}^{-1} \{v_f(S \cup \{i\}) - v_f(S)\}\\
        &= \frac{1}{\mid D \mid} \sum_{S \subseteq D \setminus \{i, j\}} {\binom{\mid D \mid -1}{|S|}}^{-1} \{v_f(S \cup \{i\}) - v_f(S)\} \\&\quad\quad\quad\quad+
        \frac{1}{\mid D \mid } \sum_{S \subseteq D \setminus \{i, j\}} {\binom{\mid D \mid -1}{|S| + 1}}^{-1} \{v_f(S \cup \{i, j\}) - v_f(S \cup \{j\})\}\\
        &= \frac{1}{\mid D \mid} \sum_{S \subseteq D \setminus \{i, j\}} {\binom{\mid D \mid-1}{|S|}}^{-1} \{v_f(S \cup \{j\}) - v_f(S)\} \\&\quad\quad\quad\quad+ 
        \frac{1}{\mid D \mid} \sum_{S \subseteq D \setminus \{i, j\}} {\binom{\mid D \mid-1}{|S + 1|}}^{-1} \{v_f(S \cup \{i, j\}) - v_f(S \cup \{i\})\}\\
        &= \frac{1}{\mid D \mid} \sum_{S \subseteq D \setminus \{j\}} {\binom{\mid D \mid-1}{|S|}}^{-1} \{v_f(S \cup \{j\}) - v_f(S)\} = \phi_j(v_f)
    \end{align*}
     
     Causal approximation (P4) follows directly from \cite{jung2022measuring}, since the proof is on the level of the value function $v_f$ the proof still holds for our global value function and our specific sampling procedure.
\qed
\end{proof}

By taking into account the causal structure of the features and the guarantees that Theorem 1 provides, we develop a method that also respects the properties of the additive global feature importance class. By interpreting the global importance of a feature by its causal contribution it has to the predictive performance our method closes a gap in explainability. Answering the questions to what extent a feature is a cause for an ML model to have a good performance differs significantly from previous Shapley-based explanability methods as they either do not measure causal contributions (\cite{covert2020feature,lundberg2017unified}) or they do not measure the contribution to the predictive performance (\cite{heskes2020causal,jung2022measuring}). Additionally, it is worth mentioning that \emph{uniqueness}  directly follows from the fact that it is based on Shapley value, and since it satisfies soundness (as showed in \cite{jung2022measuring}).

\subsection{Computing Causal Shapley Values}%Practical Implementation}
Calculating the Shapley values for each feature $X_i$ presents some practical challenges: (1) To compute the post-interventional distributions with our method the causal structure of the dataset must be known. (2) The post-interventional distribution must be transformed in an observational distribution to sample from it. (3) Computing exact Shapley values is a problem with exponential runtime because there are an exponential number of subsets $S$ of features $D$ over which we must iterate. In this Section, we introduce a pragmatic approach to handle these challenges to develop our global explanation method.

%We combine the three challenges and the solutions presented into our global explanation method, which we implemented building on top of the framework of Covert et al. \cite{covert2020understanding}.

\paragraph{Prior Knowledge on Causal Graphs.} A main assumption of computing causal Shapley values is that the causal structure is provided. This is a serious prerequisite since structural causal discovery is a challenging task itself. There are algorithms that are able to infer a structural causal model from data \cite{vowels2022d} and experiment-based approaches that identify the causal structure, %\cite{spirtes2013causal, ramsey2015scaling}
but it is hardly realistic to infer a fully specified model with all possible confounders in general \cite{shen2020challenges}. To alleviate this issue, we use causal chain graphs to calculate the feature importance. This allows us to have fewer assumptions on the causal structure of the features whenever the structure is partly unknown. %To use interventions we assume that the causal structure is known about the data. This is a serious prerequisite since structural causal discovery is a challenge itself and a lot of research has been done in recent years. There are algorithms that are able to infer a structural causal model from observational data \cite{spirtes2013causal, ramsey2015scaling}. But it is hardly realistic to infer a fully specified model with all possible confounders \cite{shen2020challenges}. We will use the notion of causal chain graphs \cite{lauritzen1989graphical} to solve this problem. Causal chain graphs represent a partial causal ordering of variables whose causal relationships are not fully known, by merging them into chain components.  We can make assumptions about the variables that are in the same component, whether they have a common confounder or whether there is mutual interaction between them. An example is given in Appendix \textit{B}. In this work, we will assume that we are given at least a causal chain graph model of the data.

\paragraph{Sampling Out-Coalition Features.} The second challenge is to estimate $\mathbb{E}[f(\mathbf{X}) | do(\mathbf{X}_S = \mathbf{x}_S)]$ by sampling from $P(\mathbf{X}_{\bar{S}} | do(\mathbf{X}_S = \mathbf{x}_S))$ according to Equation \eqref{marginalization}, since we assume no interventional data. % transform the expression $\mathbb{E}[f(X) | do(X_S = x_s)]$ in Equation (\ref{causal_v})  into a mathematical expression from which it is possible to sample. We can rewrite the expression $\mathbb{E}[f(X) | do(X_S = x_s)]$ as $\int P(X_{\bar{S}} | do(X_S = x_S)) \cdot f(X_{\bar{S}}, X_S) dx_{\bar{S}}  The rules of do-calculus \cite{pearl2012calculus} allow a mapping between interventional and observational distributions when certain conditions are satisfied in the causal graph. These rules can be applied to causal chain graph models, allowing the representation of the post-interventional distribution terms of observational distributions as follows \cite{lauritzen2002chain,heskes2020causal}: %\cite{lauritzen2002chain} applied these rules to causal chain graph models from which an interventional formula can be inferred from the observational distribution as follows \cite{heskes2020causal}:
%\begin{equation} \label{int_fullorder}
%    P(X_{\bar{S}} | do(X_S = x_S)) = \prod_{j \in \bar{S}} P(X_j | X_{Pa_j \cap \bar{S}},  X_{Pa_j \cap S})
%\end{equation}
%Note that \eqref{int_fullorder} is only applicable if we know the fully specified ordering of the variables but this is not feasible in most of the cases. 
To tackle this challenge we resort to the factorization of the post-interventional distribution for causal chain graphs \cite{heskes2020causal}:
\begin{multline} \label{int_partial}
    P(\mathbf{X}_{\bar{S}} | do(\mathbf{X}_S = \mathbf{x}_S)) = \prod_{\tau \in T_{confounding}} P(\mathbf{X}_{\tau \cap \bar{S}} | \mathbf{X}_{Pa_j \cap \bar{S}},  \mathbf{X}_{Pa_j \cap S})\\
    \times \prod_{\tau \in T_{\overline{confounding}}} P(\mathbf{X}_{\tau \cap \bar{S}} | \mathbf{X}_{Pa_j \cap \bar{S}},  \mathbf{X}_{Pa_j \cap S}, \mathbf{X}_{\tau \cap S})
\end{multline}

Equation (\ref{int_partial}) makes the distinction between confounded and not confounded chain graph components $\tau$. For chain components with confounded variables, the first part of the Equation is used. If the dependencies in a component are only due to mutual interactions between the variables the second part should be used. In contrast to Equation \eqref{marginalization}, we can now use discrete marginalization, since the causal graphs fulfill the Markov conditions.   %A more detailed explanation and derivation of the intervention formulas are given in \cite{heskes2020causal}.

\paragraph{Approximation algorithm. }  %Similar to the approximation algorithm of \cite{vstrumbelj2014explaining},  \cite{covert2020understanding} show that by their procedure the approximation algorithm will converge to the correct $\phi$ values. 
 Lastly, we combine the above insights to derive an algorithm that computes causality-aware Shapley additive importance measures for global explanations. Structurally, our algorithm follows the commonly used approach to approximate global importance values \cite{covert2020understanding}, which guarantees that the algorithm converges to the true $\phi_i$ values in the limit. The novelty of our algorithm lies in the way we sample the out-coalition features. Instead of sampling them from the conditional, observational distribution, we sample values that adhere to the causal structure of the data as described above. %Since the approximation is not influenced by the specific sampling procedure, our approach will still converge to the correct $\phi_i$. 
 The pseudocode of the overall algorithm can be seen in Algorithm \ref{algo:algo}. % the general structure introduced by \cite{covert2020understanding} with the difference that we incorporate a causality-aware sampling procedure following the results from \cite{heskes2020causal}.
%The algorithm runs through an inner loop (inner samples $m$) and an outer loop (outer samples $n$). Like most \textit{Shapley value}-based global methods, 

The algorithm requires various inputs including the dataset, a partial causal order represented as a causal chain graph, and information about confounded components or interactions in the chain graph. The algorithm works in such a way that it computes the average of many samples (Line 2) of the expression $\mathcal{L}(\mathbb{E}[f(\mathbf{X}) | do(\mathbf{X}_{S \cup i} = \mathbf{x}_{S \cup i})], Y) - \mathcal{L}(\mathbb{E}[f(\mathbf{X}) | do(\mathbf{X}_{S} = \mathbf{x}_{S})], Y)$ (Line 22) which corresponds to $v(S \cup {i}) - v(S)$ in Equation (\ref{shapley}). 

During each iteration, a data instance and a feature permutation are randomly chosen (Line 3), initiating the additive process (Line 6). This process involves incrementally adding the next feature $j$ of the permutation to the in-coalition features $S$ (Line 7). The pivotal CAGE causal sampling procedure commences in Lines 9 and 10, where a batch of size $M$ is drawn, followed by iterating over each component of the causal chain graph $\tau$ in their causal ordering. $|T|$ denotes the number of components in the causal chain graph. 

If the features in a component $\tau$ are confounded then each data point $x_{l}^{m}$ can be drawn independently (Line 13). If all feature dependencies in a chain component are induced by mutual interaction we use Gibbs sampling \cite{gelfand2000gibbs} to draw the features $\mathbf{x}^{m}_{\tau_{t}\cap \bar{S}}$ (Line 16). All sampled missing feature values $\mathbf{x}^{m}_{\bar{s}}$ are then used in Line 19 for prediction. These sampling procedures were introduced in Equation (\ref{int_partial}) and proved by \cite{heskes2020causal}. The impact of the additional feature $j$ on the predictive performance represented by the difference in the loss with and without feature $j$ (Line 21, 22) is then added to the cumulative CAGE value in Line 23.

\begin{algorithm}
    \BlankLine
    \KwIn{$\text{data } \{\mathbf{x^k}, y^k\}^K_{k=0}$, model $f$, loss $\mathcal{L}$, $N$ (outer samples), $M$ (inner samples), causal chain graph $G$, confounding, feature indicies $D$ with dimension $d$}
    \KwOut{shapley values $\frac{\phi_1}{N}, \ldots , \frac{\phi_d}{N}$}
    \BlankLine
    
    $\phi_1 = \ldots = \phi_d = 0$
    
    %margPred = $\frac{1}{N}\sum_{i=1}^{N} f(x_i)$

    \For{$i=1$ to $N$}{
        Sample $(\mathbf{x}, y)$ from $\{\mathbf{x^k}, y^k\}^K_{k=0}$ and permutation $\pi$ of D
        
        S = $\emptyset$
        
        $lossPrev = \mathcal{L}(\frac{1}{K}\sum_{k=1}^{K} f(\mathbf{x^k}), y)$

        \For{$j=1$ to $d$}{
            $S = S \cup \{\pi[j]\}$ 
            
            $\hat{y}=0$

            \For{$m=1$ to $M$}{
                \For{$t=1$ to $|T|$}{
                    \eIf{confounding($\tau_t$)}{
                        \For{$l \in \tau_t \cap \bar{S}$}{
                            $x_{l}^{m} \sim P(X_{l} | \mathbf{X}_{Pa_t \cap \bar{S}},  \mathbf{X}_{Pa_t \cap S})$
                        }
                    }
                    {
                        $\mathbf{x}^{m}_{\tau_{t}\cap \bar{S}} \sim P(\mathbf{X}_{\tau \cap \bar{S}} | \mathbf{X}_{Pa_t \cap \bar{S}},  \mathbf{X}_{Pa_t \cap S}, \mathbf{X}_{\tau \cap S})$
                    }
                }
                $\hat{y} = \hat{y} + f(\mathbf{x}_s, \mathbf{x}^{m}_{\bar{s}})$
            }
            %$\bar{y} = \frac{y}{m}$
            
            $loss = l(\frac{\hat{y}}{M}, y)$
            
            $\Delta = lossPrev - loss$
            
            $\phi_\pi[j] = \phi_\pi[j] + \Delta$
            
            $lossPrev = loss$
        }
    }
    \BlankLine    
    \KwRet{$\frac{\phi_1}{N}, ..., \frac{\phi_d}{N}$}

    \hrulefill
\caption{Approximation algorithm for CAGE values.}
\label{algo:algo}
\end{algorithm}

\section{Experiments} 

To evaluate our causal explanation framework and to compare it with other approaches, we will conduct several experiments. First, we will perform experiments on synthetic datasets. Then, we will apply our framework to a real-world example. We compare our global causality-aware explanation framework with the existing global explanation method \textit{SAGE}. 

\subsection{Experiments on Synthetic Data}\label{synth_exp}
\paragraph{Experimental Setup.} To ensure that we can assess which features are most important we conducted experiments with synthetic datasets. For these datasets, the data-generating process including its causal structure is completely known. We created three datasets with different causal structures. The first dataset only consists of independent direct causes that have the same influence on the target variable. The second dataset is of Markovian nature in the sense that we have one variable that is completely determined by its parents. Therefore, the variable is conditionally independent of its non-descendants, given its parents \cite{geiger1990logic}.  The third dataset is a mixed model consisting of both causal structures from above. This means there are causal dependencies between some variables but also direct independent variables. All three datasets are generated from structural causal models where the variables are sampled as linear combinations of the parents and pairwise independent noise terms. The exact specification of the SCMs can be found in Appendix \ref{apx:datasampling}. The graphs in Figures \ref{fig:graphs_direct} - \ref{fig:graphs_mix}  show the causal graphs induced by the SCMs. Furthermore, they show the topological ordering translated into chain graphs as the causal knowledge for our experiments.

For each dataset, we fit a linear regression model and a simple multi-layer perceptron (MLP). To implement both model types we used the \textit{scikit-learn} library with default values. Then we apply conventional SAGE and our causality-aware global explanation framework and compare the explanations.

\paragraph{Results.}

\begin{figure}   
    \begin{subfigure}{0.32\textwidth}
    \includegraphics[width=\textwidth]{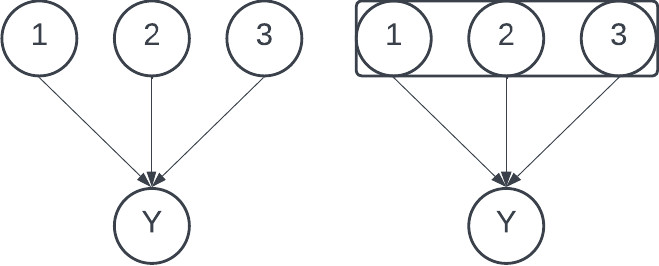}
    \caption{}
    \label{fig:graphs_direct}
    \end{subfigure}    \hfill
    \begin{subfigure}{0.32\textwidth}
\includegraphics[width=\textwidth]{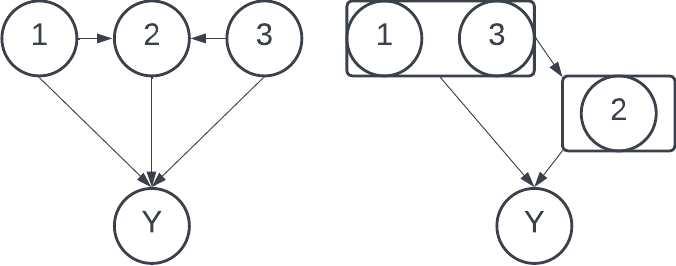}
\caption{}
\label{fig:graphs_markov}
    \end{subfigure}\hfill    
    \begin{subfigure}{0.32\textwidth}
    \includegraphics[width=\textwidth]{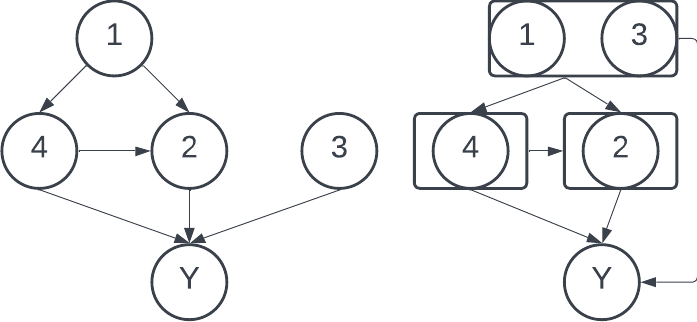}
    \caption{}
    \label{fig:graphs_mix}
    \end{subfigure}\hfill \\
    
     \begin{subfigure}{0.32\textwidth}
     \includegraphics[width=\textwidth]{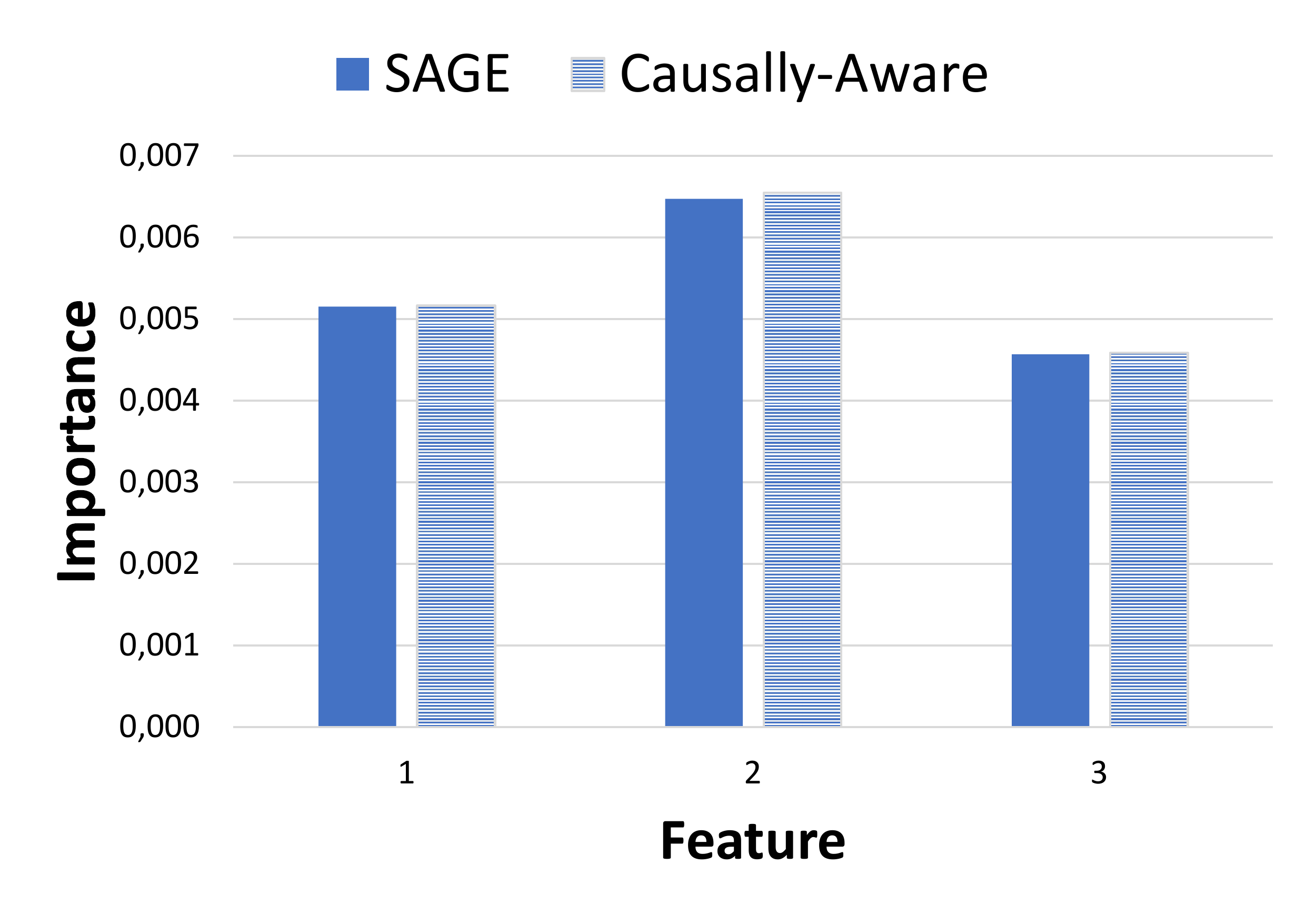}
     \caption{}
     \label{fig:results_reg_direct}
     \end{subfigure} \hfill     
    \begin{subfigure}{0.32\textwidth}
    \includegraphics[width=\textwidth]{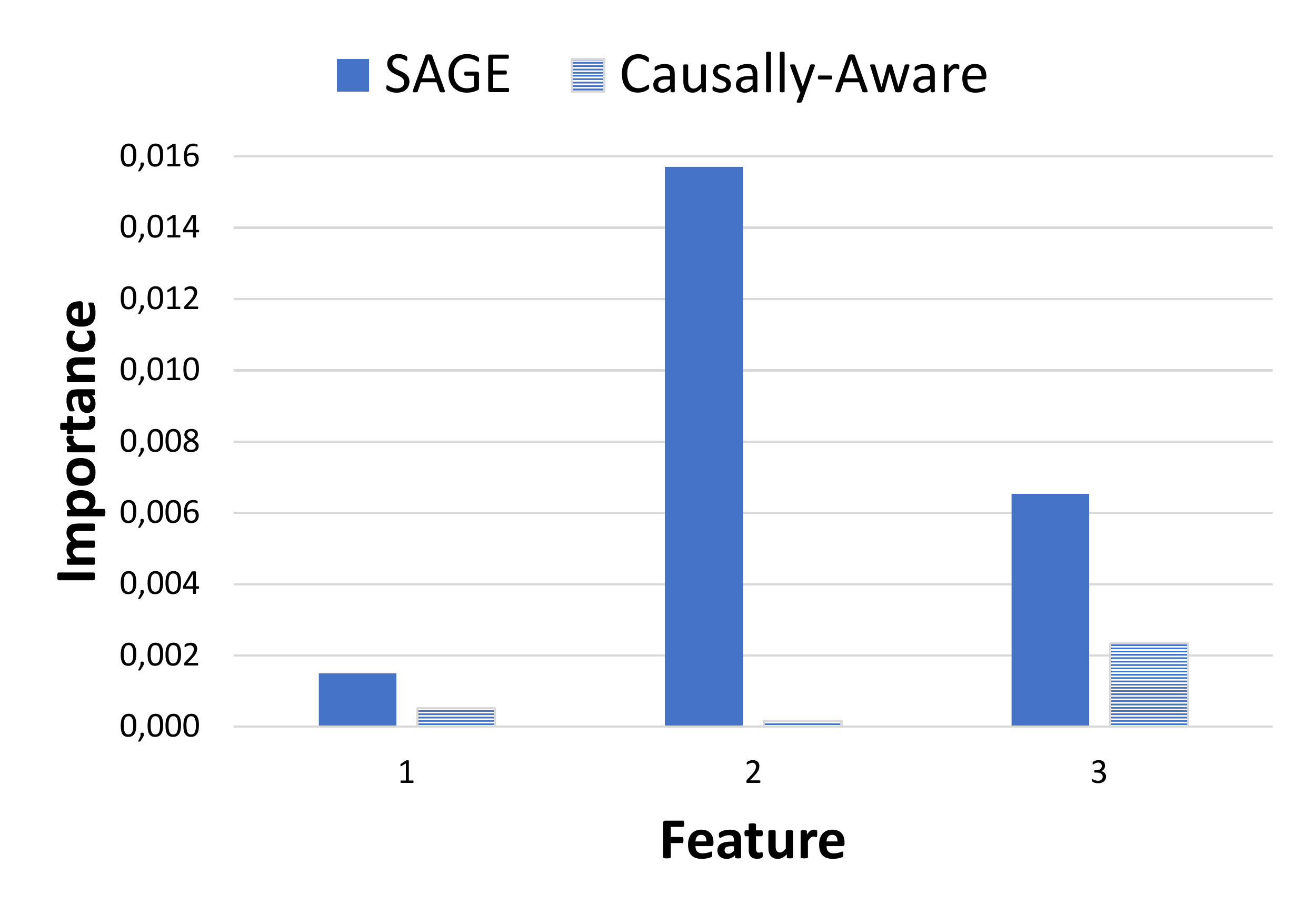}
    \caption{}
    \label{fig:results_reg_markov}
    \end{subfigure} \hfill    
    \begin{subfigure}{0.32\textwidth}
    \includegraphics[width=\textwidth]{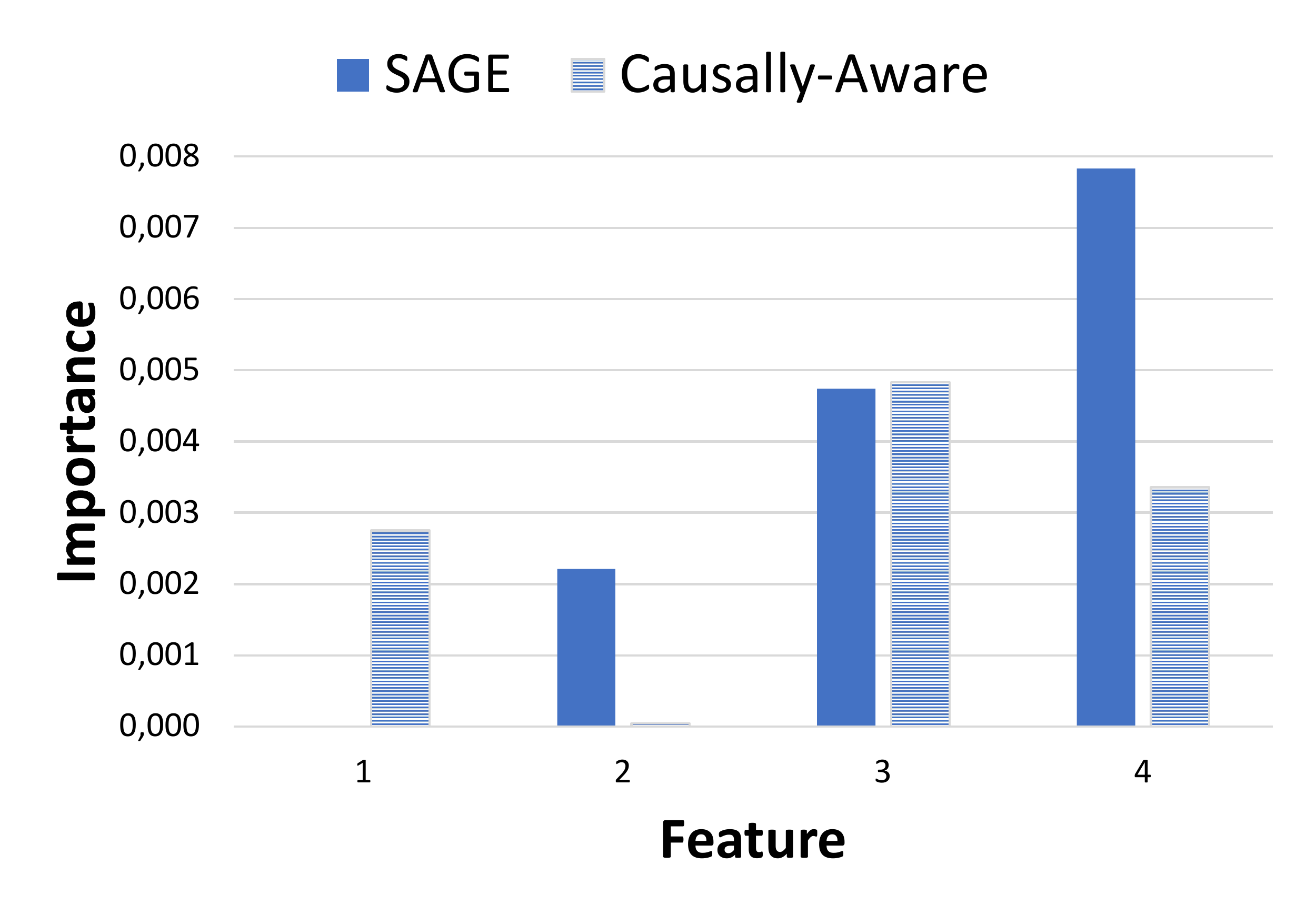}
    \caption{}
    \label{fig:results_reg_mix}
    \end{subfigure}\hfill \\
    
    \begin{subfigure}{0.32\textwidth}
    \includegraphics[width=\textwidth]{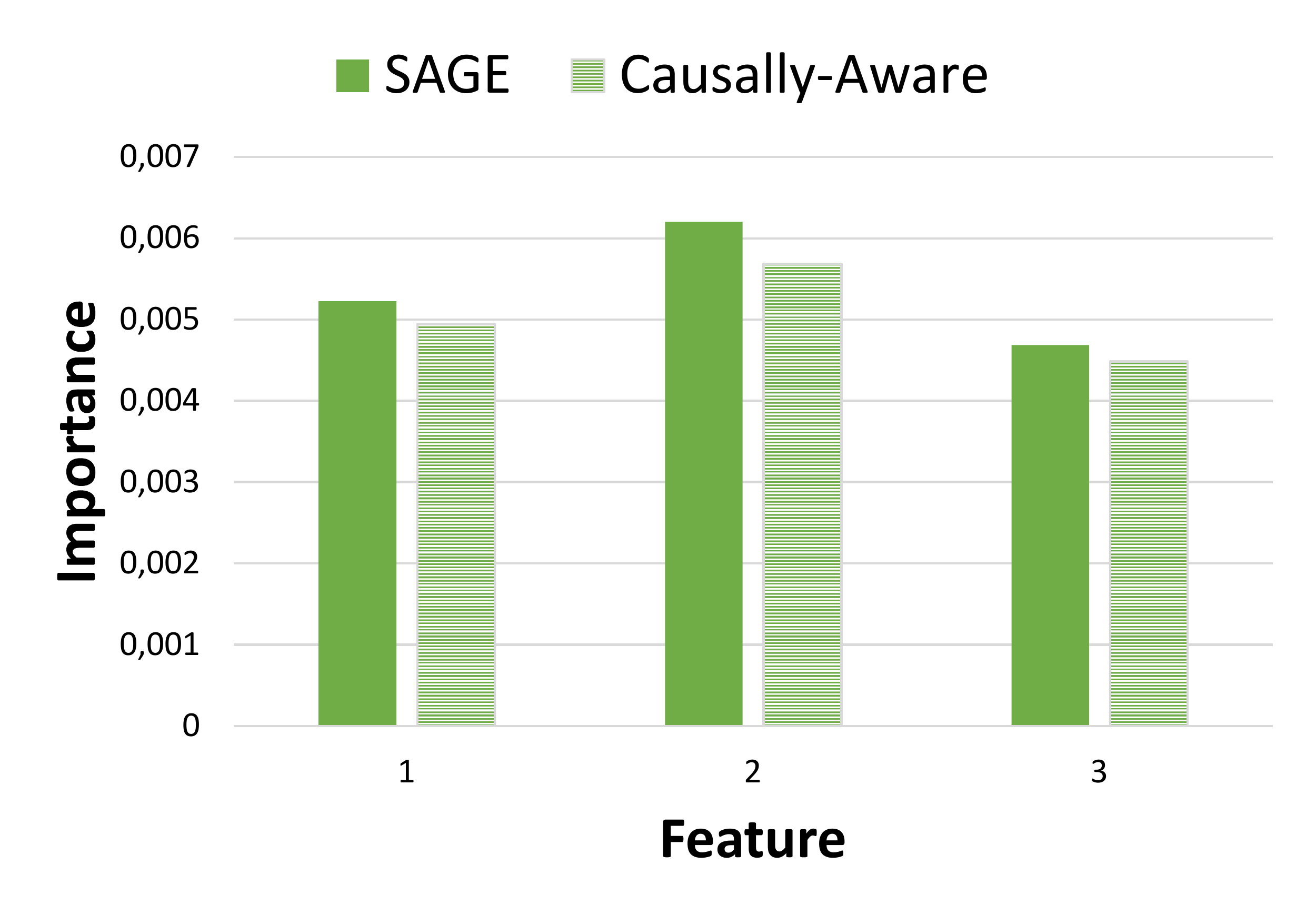}
    \caption{}
    \label{fig:results_mlp_direct}
    \end{subfigure} \hfill    
    \begin{subfigure}{0.32\textwidth}
    \includegraphics[width=\textwidth]{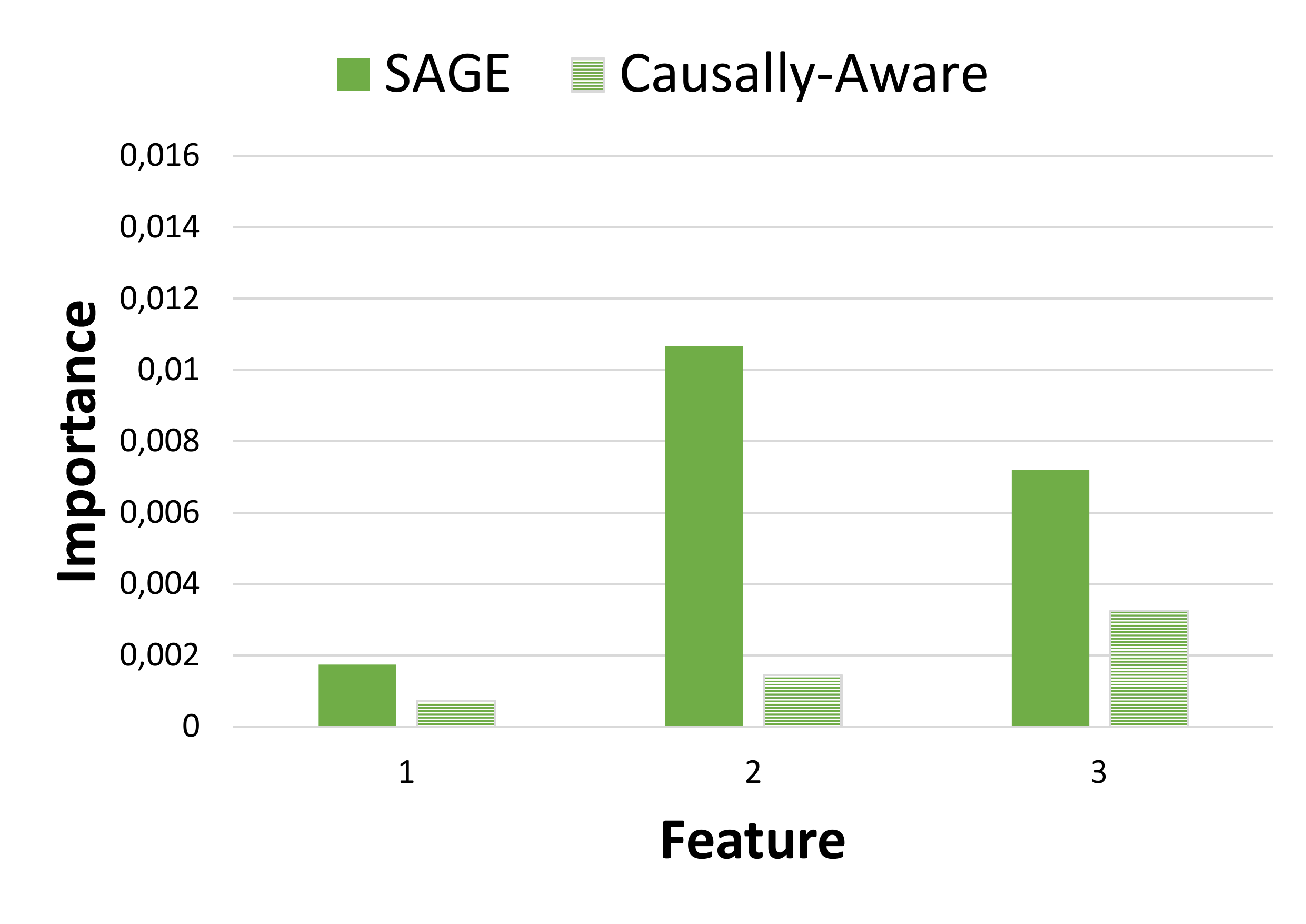}
    \caption{}
    \label{fig:results_mlp_markov}
    \end{subfigure} \hfill    
    \begin{subfigure}{0.32\textwidth}
    \includegraphics[width=\textwidth]{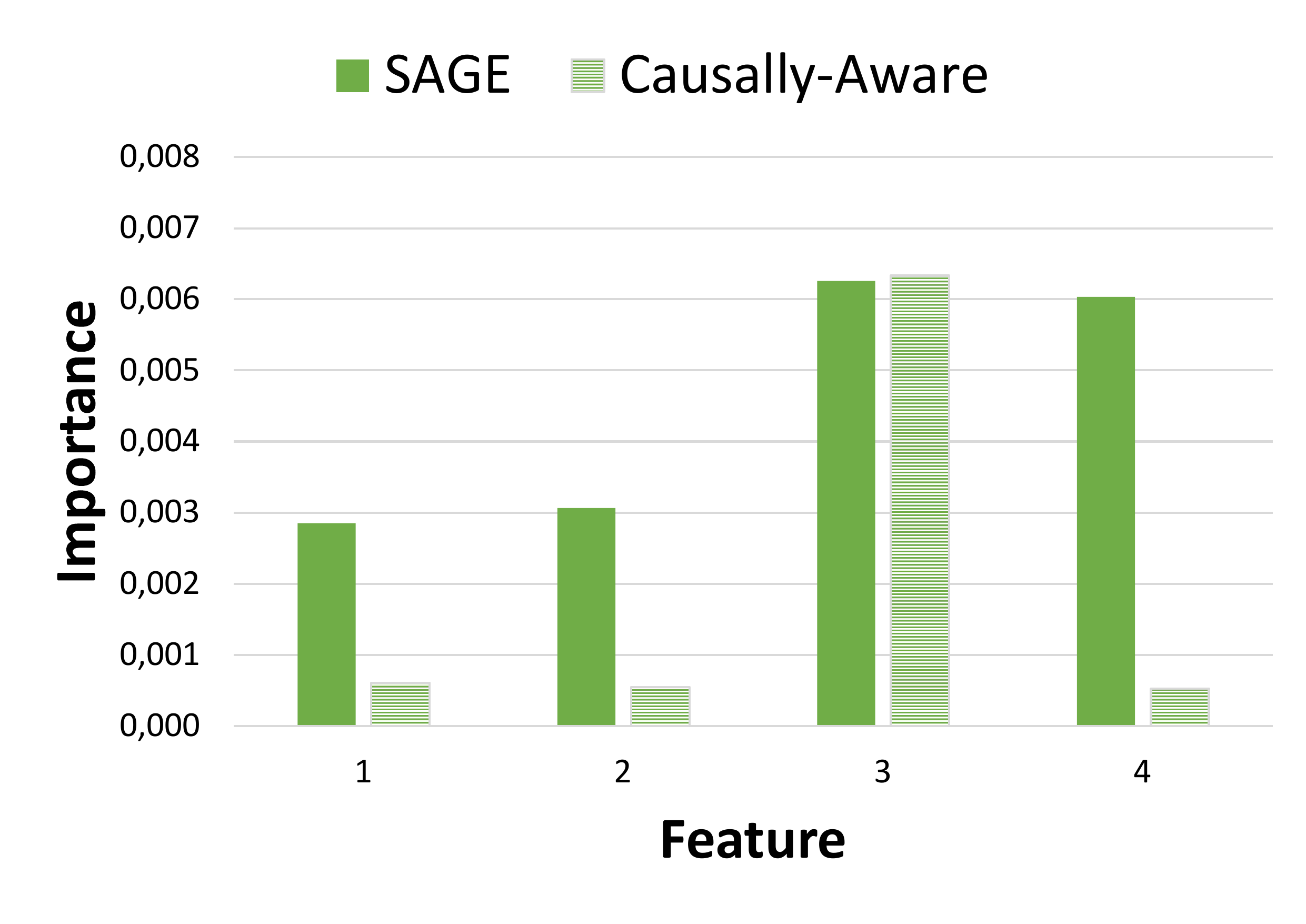}
    \caption{}
    \label{fig:results_mlp_mix}
    \end{subfigure}
    \caption{Results and data-generating causal structures for our experiments. The first row (Figures \ref{fig:graphs_direct}, \ref{fig:graphs_markov}, \ref{fig:graphs_mix}) show the true causal structure (left) of the data-generating SCMs and the corresponding causal chain graphs we use for the explanation (right). The second row (Figrues \ref{fig:results_reg_direct}, \ref{fig:results_reg_markov}, \ref{fig:results_reg_mix}, blue) shows the importance values determined for the linear regression models that were trained and evaluated on the causal structures above. The third row (Figures \ref{fig:results_mlp_direct}, \ref{fig:results_mlp_markov}, \ref{fig:results_mlp_mix}, green) shows the same information, but for the MLP models. The solid bars show values coming from SAGE, and the striped bars show values of our causally-aware global explanation method.}
    \label{fig:synth_results}
\end{figure}
%TODO fix references
Figure \ref{fig:synth_results} shows the explanatory results of the causal structures for both methods, the linear regression model (Figures \ref{fig:results_reg_direct}, \ref{fig:results_reg_markov}, \ref{fig:results_reg_mix}) and the MLP (Figures \ref{fig:results_mlp_direct}, \ref{fig:results_mlp_markov}, \ref{fig:results_mlp_mix}). In the plots, the striped bars depict our causal explanation method the solid bars depict the results when applying SAGE.

SAGE explains the feature importance for the linear regression models solely on how the target variable is built. Variables that have the highest coefficient in the deterministic function of Y get the highest importance. In contrast, our causally aware framework takes the causal structure into account. It assigns the importance of features based on their causal contribution to the target variable. For example in the Markovian dataset, even though variable 2 has the highest coefficient in the linear model of Y it gets assigned the lowest importance score because it can be completely explained by variables 1 and 3. This also applies to the Mixed-data structure. If there are independent features that do not have any causal relation with other features then the importance of that feature is the same for both explanation methods.

In general, the following characteristics can be observed in the explanations for the linear regression models:  First, if there is a variable that can be completely explained by other variables, i.e. the causal structure is clear, then this variable does not get any importance. This is in line with the causal irrelevance property introduced in Section \ref{sec:properties}. Second, if there are independent variables that are direct causes of the target then these variables have the same importance in both frameworks. Third, variables that are causes and effects at the same time get a reduction in importance but not a total deletion of importance.

The explanations of the MLP show a more nuanced picture. For the Markovian dataset, SAGE assigns feature importance similar to the linear regression model but for the mixed model dataset, it does not assign importance according to the linear model of Y. Our causally-aware method also shows different feature importance compared to the linear regression model. Features that can be completely explained by other features receive reduced feature importance but not a complete deletion of importance. For example, in the Markovian data experiment variable 2 gets reduced in importance but still gets assigned some importance, even though variable 2 is just a linear combination of variable 1 and 3. Furthermore, in the mixed-model data variables which are effects of other variables are reduced in importance. However, the root cause, variable 1, only gets small importance even though it has a high impact on the target by being the cause of variables 2 and 4. Nevertheless, the experiment with independent features shows the same results. We can see that the MLP is only able to use some of the causal information that is provided but the relationship of using the causal structure for explanation is weakened.

\subsection{Explanations on Alzheimer Data} \label{adni_exp}
Following the promising results of the synthetic datasets, in this Section we explore the distinctions between the explanations SAGE derives and our causality-aware explanations when applied to a real-world dataset.

\paragraph{Data.} To apply our framework to a real-world dataset it is necessary that we know the causal structure or at least a partial causal ordering of the features. For this experiment, we chose the Alzheimer’s Disease Neuroimaging Initiative (ADNI) dataset (\url{adni.loni.usc.edu}) because the causal structure has been investigated for this dataset \cite{jack2008alzheimer}. %Data used in the experiment of this work were obtained from the Alzheimer’s Disease Neuroimaging Initiative (ADNI) database (\url{adni.loni.usc.edu}). 
The ADNI collects data from researchers to investigate the progression of Alzheimer's disease.  The data includes MRI images, genetics, cognitive tests, %CSF,
and biomarkers as predictors of the disease. For this experiment, we do not use all features that the dataset offers but only the biomarkers fludeoxyglucose (\textit{FDG}), amyloid beta (\textit{ABETA}), phosphorylated tau (\textit{PTAU}), and the number of apolipoprotein alleles (\textit{APOE4}) additionally the age, gender, and education level as features, for simplified analysis are added. 

\paragraph{Experimental Setup.} We define a binary classification problem to use these seven features and predict if a person has Alzheimer's or not. For this classification problem, we build two models. First a simple multilayer perceptron (MLP) with five layers (\textit{layer sizes= 64, 128, 128, 64, 32}) and \textit{Adam} optimization. Second a Random Forest with 200 trees. After cleaning and normalizing the dataset it consists of 1500 instances which we split into 25\% test and 75\% training sets (Details on data handling can be found in the code repository). The trained models have an accuracy of 85\% for the MLP and 88\% for the Random Forest on the test set.

\begin{figure}
    \centering
    \includegraphics[width=0.6\textwidth]{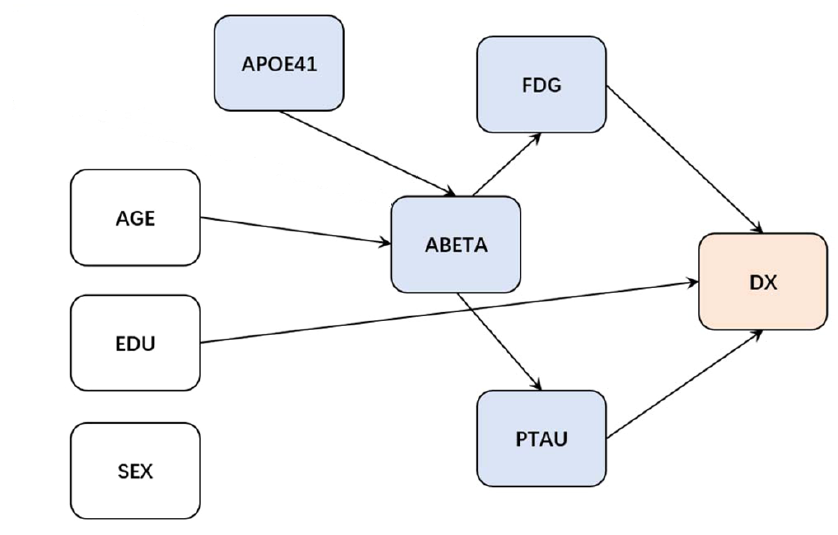}
    \caption{Gold Standard Graph from \cite{shen2020challenges}. The gold standard graph shows the causal relations between the seven features and the binary target variable \textit{DX}. Blue nodes are biomarkers and white nodes are personal information about patients. From that, we derive the causal chain graph in Figure  (\ref{fig:causalchaingraph}).}
    \label{fig:goldgraph}
\end{figure}

We analyze the causal structure of the features by referring to the research of \cite{shen2020challenges}. This study contrasts various causal structure discovery (CSD) algorithms and compares the resulting structures to a gold standard graph shown in Figure \ref{fig:goldgraph}. For our experiment, we use this gold standard graph which is based on biological and medical studies on Alzheimer's risk factors. Based on the gold standard, we define the partial causal ordering from this graph as the topological ordering: [(\textit{AGE, EDU, SEX, APOE4}), (\textit{ABETA}), (\textit{FDG, PTAU})], to show the effectiveness of our approach when only the partial causal structure is provided. Additionally, we assume confounding in the first and the third chain graph components. The resulting causal chain graph for our experiment is shown in Figure \ref{fig:causalchaingraph}.

\paragraph{Results.}

\begin{figure}
\centering
\begin{subfigure}{0.49\textwidth}
    \includegraphics[width=\textwidth]{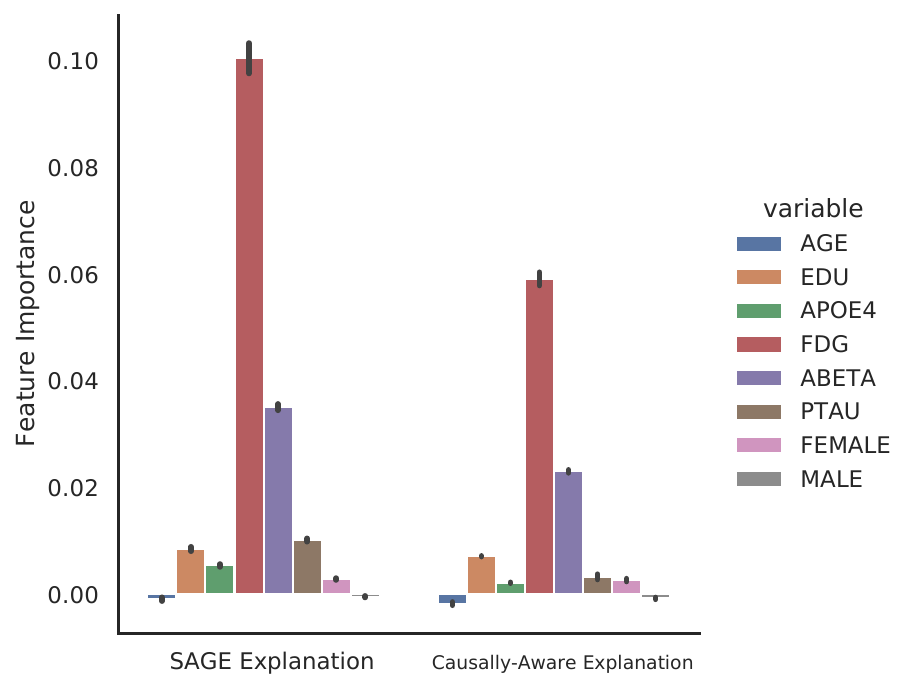}
    \caption{Multi-layer perception (MLP)}
    \label{fig:res_mlp}
\end{subfigure}\hfill
\begin{subfigure}{0.49\textwidth}
\includegraphics[width=\textwidth]{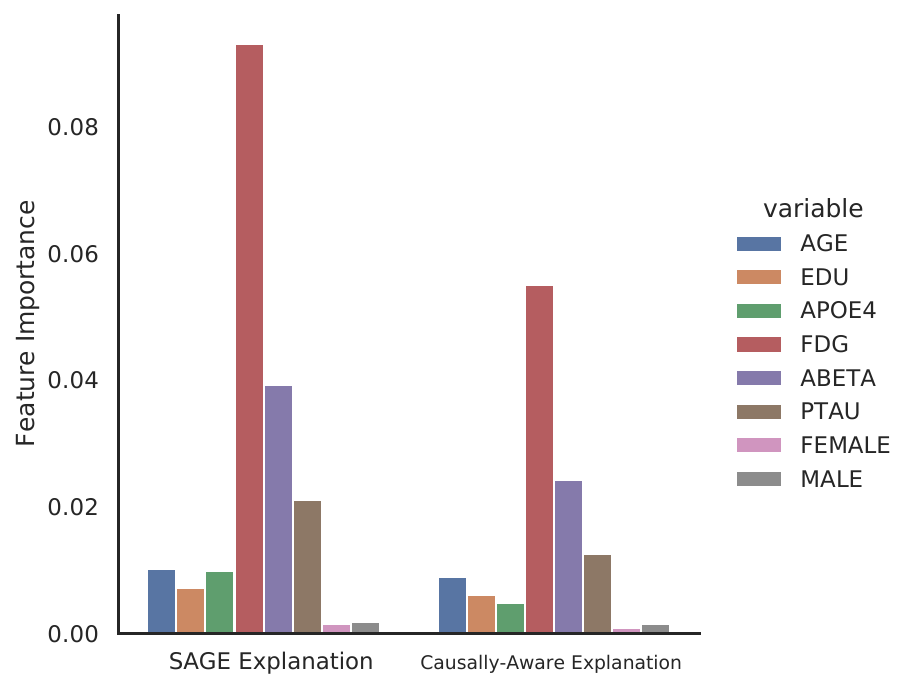}
\caption{Random Forest}
\label{fig:res_rf}
\end{subfigure}

\caption{Importance values of the ADNI data experiment. The left plot shows the feature importances for the MLP model and the right plot the feature importance of Random Forest. For each plot, the left set of bars shows the importance determined by SAGE and the right bars show the importances for our causality-aware global explanation method.}
\label{fig:res_adni}
\end{figure}

Figure \ref{fig:res_adni} depicts, the results of the experiment with the MLP (Figure \ref{fig:res_mlp}),  and the results of the Random Forest (Figure \ref{fig:res_rf}). The left bars show the explanation of \textit{SAGE} and the right bars the explanation of our causality-aware explanation framework. When comparing the two ML methods, only minor differences can be observed. More specifically, there are two dissimilarities in which the models have different feature importance rankings. These two are the feature \textit{EDU} (education) and the feature \textit{APOE4}. Both are more important for the Random Forest classifier. There are other differences in the importance but they do not have an effect on the importance ranking. One noticeable is that the importance of \textit{PTAU} is higher for the Random Forest. 

When comparing the two explanatory methods the differences are minor. There are not as extreme differences in the ranking as there are with the synthetic data. However, there are mentionable changes. For example, we observe that the importance of features that are effects of other features, e.g. \textit{PTAU} and \textit{FDG} are reduced more compared to other features in the causality-aware explanations. 

In Section \ref{sec:discussion}, we provide a detailed analysis and discussion of the possible reasons for the observed differences between the frameworks and discuss the results in more detail.

%The following differences in the explanation frameworks can be observed: In the analysis of the MLP, \textit{SAGE} assigns the highest importance to the feature \textit{FDG} also the causal explanation assigns the highest importance to this feature but with a lower value. The feature \textit{PTAU} is in the same chain component as \textit{FDG}. \textit{PTAU} is also reduced in importance in the causal explanation method. Feature \textit{ABETA} is the second most important feature both in classic \textit{SAGE} method and in the causal method. \textit{ABETA} is the direct cause of \textit{FDG} and \textit{PTAU}. The features \textit{AGE} and \textit{SEX} do not play a role and have a small importance in both explanation frameworks. \textit{EDU} (education) has a higher importance in both frameworks and is also a direct cause of the target. According to the causal graph, \textit{APOE4} is the root cause that influences \textit{ABETA}. \textit{SAGE} assigns \textit{APOE4} a relatively small importance that is even reduced in the causal framework. For the Random Forest classifier, there are almost exactly the same trends between the explanation framework as for the MLP. But there is one significant change in importance ranking for Random Forest between the two explanatory frameworks. In the causal explanation method, \textit{EDU} (education) has a higher importance value than \textit{APOE4} because \textit{APOE4} is reduced.

\section{Related Work} \label{related_work}
\textbf{Local Explanation.} Local explanations gravitate towards elucidating individual predictions and unraveling the distinctive importance attributed to features for an individual instance \cite{ribeiro2016should,lundberg2017unified}. A noteworthy methodology leveraging local explanation is the utilization of Shapley values \cite{shapley1953value}. Innovatively adapted from game theory, Shapley values have been employed to measure feature contribution towards a model’s output, as showcased in frameworks like SHAP \cite{lundberg2017unified}. Since its inception, SHAP has evolved, adapting to a diverse array of tasks and explanatory objectives through various extensions and modifications. Notable among these are \textit{KernelSHAP} \cite{lundberg2017unified,covert2021improving}, \textit{TreeSHAP} \cite{lundberg2020local}, and \textit{LossSHAP} \cite{lundberg2020local,covert2020feature}. KernelSHAP, a model-agnostic explainable method, is appreciated for its adaptability across numerous model types. In contrast, TreeSHAP is specially tailored for tree models, offering dedicated insights into tree-based model interpretations. LossSHAP diverges in its explanatory focus. Rather than adhering to traditional approaches, it emphasizes the influence of features based on evaluation metrics. For instance, it evaluates the importance of a feature by analyzing its impact on specific evaluation criteria such as the mean squared error in regression contexts. 

\textbf{Causal Local Explanation.} An evolution in local explanations is witnessed in the integration of causality, fostering a more nuanced and reliable interpretation. Aas et al.  \cite{aas2021explaining} extend the \textit{KernelSHAP} method so that it can handle highly correlated features. Another line of studies investigates how feature dependencies and \textit{Shapley values} can be interpreted from a causal perspective. Frye et al. \cite{frye2020asymmetric} present \textit{Asymmetric Shapley values} where they incorporate causal knowledge by only allowing possible permutations of features that comply with the causal structure of the input features when computing \textit{Shapley values}. Janzing et al. \cite{janzing2020feature} tackle the question of how to deal with out-coalition features in \textit{SHAP}-based methods. They replace conditional sampling in the \textit{Shapley value} computation with conditioning by intervention with do-calculus. Similarly, Heskes et al. \cite{heskes2020causal} and Jung et al. \cite{jung2022measuring} use do-calculus to compute the causal contribution of a feature to the models' prediction. \textit{Shapley Flow} is a \textit{Shapley value}-based method that explains model predictions from a causal perspective. The authors suggest not to assign importance to variables in the causal graph but to assign importance to the edges of the causal graph \cite{wang2021shapley}. Another interpretation of causal feature importance is given by abductive explanations which generate a minimal subset of features that are sufficient for the prediction \cite{biradar2024axiomatic}. This interpretation is closely linked to the causal strength quantification notion of \cite{chockler2004responsibility}.

\textbf{Global Explanation.} Global explanations pivot towards explaining the entire model mechanism by, for instance, providing the most important features for the model to make a prediction. SAGE (Shapley Additive Global Explanations) \cite{covert2020understanding} emerges as a quintessential global explanation methodology. SAGE introduces additive importance measures as a similar class of methods like additive feature attribution methods \cite{lundberg2017unified}. In this class, the importance of a feature is defined as the predictive power that it contributes rather than the absolute effect it has on the prediction. This means that \textit{SAGE} measures if a feature makes a prediction more or less correct, according to evaluation metrics, whereas \textit{SHAP}-based methods measure the pure change that features have on the prediction. \textit{SAGE} is, therefore, the global equivalent of \textit{LossSHAP}. For an in-depth review of global XAI methods, we refer to \cite{saleem2022explaining}.

\section{Discussion} \label{sec:discussion}

In our study, we embarked on an exploration of causal global explanation methods. We hypothesized that these methods, grounded in causal foundations, assign importance to features in a manner that is more congruent with their actual causal contributions towards model predictions, as opposed to the SAGE framework which lacks a causal basis.

Our results with linear regression on synthetic datasets substantiate this hypothesis. By leveraging causal knowledge, our method assigns importance to features more intuitively, yielding explanations that closely mirror actual causal feature contributions to predict the target variable. This precision is attributable to the availability of structured causal models (SCMs), clarifying feature constructions and contributions. Remarkably, the proposed method assigns minimal importance to features that significantly contribute to the target but are the effects of other features, aligning the explanations with actual causal contributions to the predictive power.

Interestingly, our causal framework exhibited a propensity to assign significance to root causes within the SCM, even in scenarios where these root causes did not exert direct influences on the target. This observation is pivotal, aligning with human cognitive patterns in causality attribution, and echoes theories suggesting humans evaluate each event within a causal chain based on its impact on the outcome \cite{lewis1973causation,spellman1997crediting,sloman2005causal}. The crediting causality hypothesis suggests that all events in a causal chain are evaluated on how much they change the outcome. This leads to the fact that in simpler mechanisms the root cause is often given as one of the main causes \cite{spellman1997crediting}. %Another theory suggests that humans sample over possible scenarios to assign causation, this would lead to less importance to the root \cite{lewis1973causation}.

A salient observation is the causal method's tendency to attribute reduced absolute importance values compared to the traditional SAGE framework. This discrepancy may stem from algorithmic calculations and the dataset's structural composition, necessitating cautious inter-framework comparisons. Furthermore, the causal sampling method, by reducing outliers, may contribute to lower absolute importance values.

We additionally evaluated our method on a real-world example. As expected the results of this experiment were not so clear as for the synthetic examples. As a preliminary, it should be noted that Alzheimer's is still a rather unexplored disease and the underlying mechanisms are not fully understood. This is also how the gold standard graph, which is used as a basis for causal knowledge, can be classified. The graph is based on biological and medical observational analyses of the ADNI dataset in which the true causal mechanisms are not fully known. This means that the possibility of unobserved confounding needs to be assumed. %It is in the nature of things that experiments based on real data cannot be interpreted so clearly.

Nevertheless, we discuss our results for the real-world data based on the characteristics we have developed for the synthetic data. A repeating pattern that can be observed both in the synthetic data experiments and in the real-world application is that features that are solely effects of other features have a reduced feature importance. This is in line with the characteristics of our causality-aware explanation method that we put forward above. However, the concentration of feature importance on the root cause cannot be observed in real-world data experiments. We attribute this to these possible reasons: First, the model is not able to learn and use the causal structure. Models like MLP and Random Forests are high-dimensional ML models that only learn statistical correlations. Augmenting their explanation with causal knowledge does not necessarily mean that the models actually rely on it. Second, we do not provide complete causal ordering but only partial causal ordering with chain graphs. This means that some causal knowledge is lost and cannot be exploited for the explanation. Signs supporting these arguments can be observed in the explanations of the MLP with mixed-model synthetic data in Section \ref{synth_exp}. There, where the causal relations are more complicated, we observe the patterns of the real-world data application.

The analysis of the results thus reveals some interesting aspects. For explaining simple models like linear regression models the causal structure can be used to almost exactly represent the causal contributions of features. If the models become more high-dimensional this capability becomes less. Due to the fact that high-dimensional models tend to learn merely statistical correlations, we suspect that the causal information we provide is lost at the global level of explanation.

The primary challenge of our global causal explanation method lies in requiring a predefined causal structure for features, a difficult task as determining causality itself is an ongoing research area \cite{vowels2022d}. While we utilize causal chain graphs, their practicality diminishes with an increasing number of features, complicating the division into chain components. This complexity was evident in our ADNI dataset experiment, where unclear causal structures and minimal causal effect strengths made the results and their interpretations ambiguous. Users must scrutinize the causal structure's origin, effect strengths, and feature classifications in chain components for valid interpretations.

To the best of our knowledge, our work is the first one that incorporates causal knowledge into global Shapley - value-based explanation methodologies. Preliminary comparisons with local explanation methods \cite{aas2021explaining,jung2022measuring} indicate a consensus, underscoring the enhancement of explanatory accuracy and coherence when causal structures are incorporated. The proposed method demonstrated similar results and improvements, justifying our results on global-level explanations.

\section{Conclusion}
In this paper, we propose CAGE, a causality-aware global additive explanation framework based on Shapley values. We show that it is able to generate explanations that align with desirable causal properties, and outlined an algorithm for estimating its values. To this end, we introduced a novel sampling procedure for out-coalition features that respects their causal relation. Most notably, in contrast to previous global explanation approaches, our approach takes away the burden of the independence assumption among input features. Application of CAGE to both synthetic and real-world datasets shows that CAGE respects the causal relations of input features while explaining predictive models. We argue that this leads to more intuitive and faithful explanations of AI.   %In addition, we show a new interpretation of causal global feature importance, namely causal contribution to the predictive power of each feature to the model, that results from the novel sampling procedure.  
  %We confirmed these ideas with experiments that are carried out through both synthetic data and real-world data. 

In future work, causal explanation methods based on Shapley values should investigate how the strong prerequisite that the causal structure of the features must be given can be overcome. This is the basis for the widespread use of causal explanation methods. Important points of orientation for this could be studies that investigate causal reasoning and causal learning under uncertainty or partially confounded settings.

%that is able to determine the importance of each feature in an ML model. Research has shown that Shapley value methods for explainability that assume independence among features make misleading explanations. Our method uses known causal relationships between features and takes these into account when assessing the importance of features. We evaluated the causal explanation method on synthetic datasets and also on a real-world application. Results of synthetic experiments show that the causal explanation method is closer to the real contribution of the features to the target. The key takeaways of this paper are that incorporating causal knowledge into global explanation methods leads to better explanations of feature importance. Beyond that, the analysis shows that causality is not just an option that can be considered in the explanation or not but rather a necessity and the basis from which the development of an explanatory method is to be started. Only then the resulting explanations can meet the philosophical and psychological definition of an explanation.

% Acknowledgments---Will not appear in anonymized version
%\acks{We thank a bunch of people and funding agency.}

\section*{Acknowledgements and Funding}
Nils Ole Breuer works in the gemeinwohlorientierter KI-Anwendungen (Go-KI) project (Offenes Innovationslabor
KI zur Förderung gemeinwohlorientierter KI-Anwendungen), funded by the German Federal Ministry of Labour and Social Affairs
(BMAS) under the funding reference number DKI.00.00032.21.

This research was partially funded by the Hybrid Intelligence Center, a 10-year programme funded by the Dutch Ministry of Education, Culture and Science through the Netherlands Organisation for Scientific Research, \url{https://hybridintelligence-centre.nl}, grant number 024.004.022.

\newpage
\bibliographystyle{splncs04}
\bibliography{main.bib}

\appendix

\section{Data - Generating Causal Models} \label{apx:datasampling}

These are the structural causal models used for the data-generation process of the synthetic data experiments used in Section \ref{synth_exp}.

\subsection{Direct-Cause structure}
The SCM that induced the graph in Figure \ref{fig:graphs_direct}:
\begin{align*}
    N &= \mathcal{N}(0, 1)\\
    X_1 &= \mathcal{N}(0, 1)\\
    X_3 &= \mathcal{N}(0, 1)\\
    X_2 &= \mathcal{N}(0, 1)\\
    Y &= X_1 + X_2 + X_3 + N_Y\\
\end{align*}

\subsection{Markovian Structure}
The SCM that induced the graph in Figure \ref{fig:graphs_markov}:
\begin{align*}
    N &= \mathcal{N}(0, 1)\\
    X_1 &= \mathcal{N}(1.5, 1)\\
    X_3 &= \mathcal{N}(0.5, 2)\\
    X_2 &= X_1 + X_3 + N_{V_2}\\
    Y &= X_1 + 2X_2 + X_3 + N_Y\\
\end{align*}

\subsection{Mixed structure}
The SCM that induced the graph in Figure \ref{fig:graphs_mix}:
\begin{align*}
    N &= \mathcal{N}(0, 1)\\
    X_1 &= \mathcal{N}(1.5, 1)\\
    X_3 &= \mathcal{N}(0.5, 2)\\
    X_4 &= X_1 + N_{V_4}\\
    X_2 &= X_1 + X_4 + N_{V_2}\\
    Y &= 0.3X_2 + X_3 + 2X_4 + N_Y\\
\end{align*}

\end{document}